# Causal Inference by Surrogate Experiments: z-Identifiability

Elias Bareinboim and Judea Pearl
Cognitive Systems Laboratory
Department of Computer Science
University of California, Los Angeles
Los Angeles, CA. 90095
{eb,judea} at cs.ucla.edu

## Abstract

We address the problem of estimating the effect of intervening on a set of variables $X$ from experiments on a different set, $Z$, that is more accessible to manipulation. This problem, which we call $z$-identifiability, reduces to ordinary identifiability when $Z = \emptyset$ and, like the latter, can be given syntactic characterization using the *do-calculus* [Pearl, 1995; 2000]. We provide a graphical necessary and sufficient condition for $z$-identifiability for arbitrary sets $X, Z$, and $Y$ (the outcomes). We further develop a complete algorithm for computing the causal effect of $X$ on $Y$ using information provided by experiments on $Z$. Finally, we use our results to prove completeness of *do*-calculus relative to $z$-identifiability, a result that does not follow from completeness relative to ordinary identifiability.

## 1 Introduction

The relation between passive and experimental observations, and how they can aid the estimation of causal effects, is of central interest in the empirical sciences.

In this line of research, the *identification* problem (*ID*, for short) asks whether causal effects can be computed from the joint distribution P over the observed variables, and theoretical knowledge encoded in the form of a causal diagram $G$.

This problem has been extensively studied in the literature, and [Pearl, 1995; 2000] gave it rigorous mathematical treatment based on the structural semantics, and introduced several graphical conditions such as the "back-door" and "front-door" criteria, which was later generalized by his *do-calculus*. In the last decades, a number of conditions had emerged for non-parametric identifiability such as the ones given by [Spirtes, Glymour, and Scheines, 1993; Galles and Pearl, 1995; Pearl and Robins, 1995; Halpern, 1998; Kuroki and Miyakawa, 1999]. In a series of breakthrough results starting with the development of the concept of C-component [Tian and Pearl, 2002], the *do-calculus* was finally shown to be complete [Huang and Valtorta, 2006; Shpitser and Pearl, 2006]. This result implies that there exists a finite sequence of applications of the rules of do-calculus that derives the target causal effect $Q$ in terms of the observational distribution $P$ if (and only if) $Q$ is identifiable. The same work also provided algorithms that return a mapping from $P$ to $Q$ whenever $Q$ is identifiable.

In real world applications, it is not uncommon that the quantity $Q$ is unidentifiable, i.e., the distribution $P$ together with the graph $G$ are not able to unambiguously determine $Q$. A natural question arises whether the investigator could perform some auxiliary experiments (not necessary spelled out in $Q$), which would enable him/her to estimate the desired causal effects.

For instance, consider the causal diagram $G$ in Fig. 1(a). Suppose one is interested in assessing the effect $Q$ of cholesterol levels ($X$) on heart disease ($Y$), and data about subjects' diet ($Z$) is also collected. It is clear that $Q$ is unidentifiable from the assumptions embodied in $G$, but it is infeasible in reality to control subjects' cholesterol level by intervention. Assume that an experiment can be conducted in which the subjects' diet ($Z$) is randomized; a natural question emerges whether $Q$ is computable given this additional piece of experimental information?

Surprisingly, this ubiquitous problem has not received a thorough formal treatment. We introduce a variation of the *ID* problem to fill in this gap. Consider a setting in which, in addition to the information available in an ordinary *ID* instance (distribution $P$ and graph $G$), further experiments can be performed over a set of variables $Z$; decide whether the target causal effects can be computed from the available information at hand. This extension generalizes the *ID* problem (when $Z = \emptyset$ the two problems coincide) and is called here the $z$-identification problem (*zID*, for short). The $Z$ is called surrogate experiments, for obvious reasons.

Syntactically, the *zID* problem amounts to transforming $P(y|\hat{x})$[1] into an equivalent expressions in *do*-calculus such that only members of $Z$ may contain the hat symbol. Applying this rationale for the example given above (Fig. 1(a)) entails the following reduction in the *do*-calculus. First apply Rule 3 to add $\hat{z}$,

$$P(y|\hat{x}) = P(y|\hat{x}, \hat{z}) \text{ since } (Y \perp\!\!\!\perp Z|X)_{G_{\overline{XZ}}}$$

Then apply Rule 2 to exchange $\hat{x}$ with $x$:

$$P(y|\hat{x}, \hat{z}) = P(y|x, \hat{z}) \text{ since } (Y \perp\!\!\!\perp X|Z)_{G_{\underline{X}\overline{Z}}}$$

This last expression can be rewritten as,

$$P(y|x, \hat{z}) = \frac{P(y, x|\hat{z})}{P(x|\hat{z})} \quad (1)$$

This expression shows that performing an experiment on $Z$ suffices to yield "identifiability" of the causal effect of $X$ on $Y$ without experimenting over $X$. [2]

The subtlety of this problem can be illustrated by noting that in the graph in Fig. 1(a) the effect is *z*-identifiable from $P(V)$ and $P(X, Y|\hat{Z})$ in $G$, whereas in the graph in Fig. 1(b) it is not (to be shown later). The only difference between these two graphs is the bidirected edge between the pairs $(X, Z)$ and $(X, Y)$.

One might surmise that *zID* can be represented by a mutilated graph in which the edges incoming to $Z$ are cut, and the problem would then be solved as ordinary identifiability. Unfortunately, this is not the case as shown in the graph in Fig. 1(c) where $Q = P(y|\hat{x})$. The option of manipulating $Z$ does not enable us to compute the $Z$-specific causal effect of $X$ on $Y$, $P(y|\hat{x}, z)$ which, if available, would allow us to compute the overall causal effect by averaging over $Z$. Although $Q' = P(y|\hat{x}, \hat{z})$ can be established from the mutilated graph, it does not help in establishing the $Z$-specific causal effect, or $Q$.

The first formal treatment of this problem [Pearl, 1995] led to the following sufficient condition for admitting a surrogate variable $Z$ for the causal effect $P(y|\hat{x})$:

(i) $X$ intercepts all directed paths from $Z$ to $Y$, and

(ii) $P(y|\hat{x})$ is identifiable in $G_{\overline{Z}}$.

These conditions are satisfied indeed in the model of Fig. 1(a) but not in 1(b) or 1(c). Pearl's criterion is sufficient but was not shown to be necessary. Additionally, it was not extended to the case where $Z$ and $X$ are sets of variables. At the same time, the syntactic condition above, which requires the existence of a do-calculus transformation expression containing only do(z) terms is declarative, but is not computationally effective, since it does not specify the sequence of rules leading to the needed transformation, nor does it tell us if such a sequence exists. Even though do-calculus is complete for identifying causal effects, it is not immediately clear whether it is complete for *zID*.

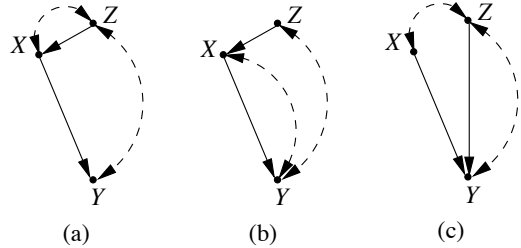

Figure 1: Causal diagrams illustrating z-identifiability of the causal effect $Q = P(y|\hat{x})$. $Q$ can be identified by experiments on $Z$ in model (a), but not in (b) and (c).

This paper provides a systematic study of *z*-identifiability building on Pearl's condition and the previous results from the identifiability literature; our contributions are as follows:

- We provide a necessary and sufficient graphical condition for the problem of *z*-identification when **Z** is a set of variables.

- We then construct a complete algorithm for deciding *z*-identification of joint causal effects and returning the correct formula whenever those effects are *z*-identifiable.

- We further show that *do*-calculus is complete for the task of *z*-identification.

## 2 Notation and Definitions

The basic semantical framework in our analysis rests on *probabilistic causal models* as defined in [Pearl, 2000, pp. 205], which are also called structural causal models or data-generating models. In the structural causal framework [Pearl, 2000, Ch. 7], actions are modifications of functional relationships, and each action $do(\mathbf{X} = \mathbf{x})$ on a causal model $M$ produces a new model $M_\mathbf{x} = \langle \mathbf{U}, \mathbf{V}, \mathbf{F}_\mathbf{x}, P(\mathbf{U}) \rangle$, where $F_\mathbf{x}$ is obtained after replacing $f_X \in \mathbf{F}$ for every $X \in \mathbf{X}$ with a new function that outputs a constant value $x$ given by $do(\mathbf{X} = \mathbf{x})$.

---

[1] We will use $P(y|\hat{x})$ interchangeably with $P_x(y)$ or $P(y|do(x))$. We will also call the interventional operator $do()$ as the "hat" operator.

[2] The expression also shows that only one level of $Z$ suffices for the identification of $P(y|\hat{x})$ for any value of $y$ and $x$. In other words, $Z$ need not be varied at all; it can simply be held constant by external means and, if the assumptions embodied in $G$ are valid, the r.h.s. of eq. (1) should attain the same value regardless of the (constant) level at which $Z$ is being held constant. In practice, however, several levels of $Z$ will be needed to ensure that enough samples are obtained for each desired value of $X$.

We follow the conventions given in [Pearl, 2000]. We will denote variables by capital letters and their values by small letters. Similarly, sets of variables will be denoted by bold capital letters, sets of values by bold letters. We will use the typical graph-theoretic terminology with the corresponding abbreviations $Pa(\mathbf{Y})_G$, $An(\mathbf{Y})_G$, and $De(\mathbf{Y})_G$, which will denote respectively the set of observable parents, ancestors, and descendants of the node set $\mathbf{Y}$ in $G$. By convention, these sets will include the arguments as well, for instance, the ancestral set $An(\mathbf{Y})_G$ will include $\mathbf{Y}$. We will usually omit the graph subscript whenever the graph in question is assumed or obvious. A graph $G_\mathbf{Y}$ will denote the induced subgraph $G$ containing nodes in $\mathbf{Y}$ and all arrows between such nodes. Finally, $G_{\overline{\mathbf{X}}\underline{\mathbf{Z}}}$ stands for the edge subgraph of $G$ where all incoming arrows into $\mathbf{X}$ and all outgoing arrows from $\mathbf{Z}$ are removed.

We build on the problem of identifiability, defined below, which expresses the requirement that causal effects must be computable from a combination of passive data $P$ and the assumptions embodied in a causal graph $G$ (*without* assuming any availability of additional experimental information).

**Definition 1** (Causal Effects Identifiability (Pearl)). *Let $\mathbf{X}, \mathbf{Y}$ be two sets of disjoint variables, and let $G$ be the causal diagram. The causal effect of an action $do(\mathbf{X} = \mathbf{x})$ on a set of variables $\mathbf{Y}$ is said to be identifiable from $P$ in $G$ if $P_\mathbf{x}(\mathbf{y})$ is (uniquely) computable from $P(V)$ in any model that induces $G$.*

The following Lemma is the operational way to prove that a causal quantity is not identifiable given the assumptions embedded in $G$.

**Lemma 1.** *Let $\mathbf{X}, \mathbf{Y}$ be two sets of disjoint variables, and let $G$ be the causal diagram. $P_\mathbf{x}(\mathbf{y})$ is not identifiable in $G$ if there exist two causal models $M^1$ and $M^2$ compatible with $G$ such that $P_1(\mathbf{V}) = P_2(\mathbf{V})$, and $P_1(\mathbf{y}|do(\mathbf{x})) \neq P_2(\mathbf{y}|do(\mathbf{x}))$.*

*Proof.* The latter inequality rules out the existence of a function from $P$ to $P_\mathbf{x}(\mathbf{y})$. □

Next, we formally introduce the problem of $z$-identifiability that generalizes the problem of identifiability whereas it is no longer assumed that experimental information is not available at all, but there exists a set of variable $\mathbf{Z}$ in which experiments were performed and now is available for use. In other words, the explicit acknowledgement of the existence of the set $\mathbf{Z}$ adds a degree of freedom for the researcher, making the analysis more flexible and perhaps realistic.

**Definition 2** (Causal Effects $z$-Identifiability). *Let $\mathbf{X}, \mathbf{Y}, \mathbf{Z}$ be disjoint sets of variables, and let $G$ be the causal diagram. The causal effect of an action $do(\mathbf{X} = \mathbf{x})$ on a set of variables $\mathbf{Y}$ is said to be $z$-identifiable from $P$ in $G$, if $P_\mathbf{x}(\mathbf{y})$ is (uniquely) computable from $P(\mathbf{V})$ together with the interventional distributions $P(\mathbf{V} \setminus \mathbf{Z}'|do(\mathbf{Z}'))$, for all $\mathbf{Z}' \subseteq \mathbf{Z}$, in any model that induces $G$.*

Armed with this new definition, we state next the sufficiency of the *do*-calculus for $zI\mathcal{D}$ that is analogous to [Pearl, 2000, Corol. 3.4.2] in respect to identification.

**Theorem 1.** *Let $\mathbf{X}, \mathbf{Y}, \mathbf{Z}$ be disjoint sets of variables, let $G$ be the causal diagram, and $Q = P(\mathbf{y}|do(\mathbf{x}))$. $Q$ is $zI\mathcal{D}$ from $P$ in $G$ if the expression $P(\mathbf{y}|do(\mathbf{x}))$ is reducible, using the rules of do-calculus, to an expression in which only elements of $\mathbf{Z}$ may appear as interventional variables.*

*Proof.* The result follows from soundness of do-calculus and the definition of $z$-identifiability. □

It is clear that if we have an efficient procedure to establish $zI\mathcal{D}$, we can immediately decide $I\mathcal{D}$ by setting $\mathbf{Z} = \emptyset$. On the other hand, to be able to establish the converse of Theorem 1, we need to understand the conditions for non-$zI\mathcal{D}$, and so, we state next the analogous of Lemma 1 in this context.

**Lemma 2.** *Let $\mathbf{X}, \mathbf{Y}, \mathbf{Z}$ be disjoint sets of variables, and let $G$ be the causal diagram. $P_\mathbf{x}(\mathbf{y})$ is not $z$-identifiable in $G$ if there exist two causal models $M^1$ and $M^2$ compatible with $G$ such that $P^1(\mathbf{V}) = P^2(\mathbf{V})$, $P^1(\mathbf{V} \setminus \mathbf{Z}'|do(\mathbf{Z}')) = P^2(\mathbf{V} \setminus \mathbf{Z}'|do(\mathbf{Z}'))$, for all $\mathbf{Z}' \subseteq \mathbf{Z}$, and $P^1_\mathbf{x}(\mathbf{y}) \neq P^2_x(\mathbf{y})$.*

*Proof.* Let $I$ be the set of interventional distributions $P(\mathbf{V}\setminus\mathbf{Z}'|do(\mathbf{Z}'))$, for any $\mathbf{Z}' \subseteq \mathbf{Z}$. The latter inequality rules out the existence of a function from $P, I$ to $P_\mathbf{x}(\mathbf{y})$. □

While Lemma 2 might appear convoluted, it is nothing more than a formalization of the statement "Q cannot be computed from information set $S$ alone." Naturally, when S has two components, $\langle P, I \rangle$, the Lemma becomes lengthy. Even though the problems of $I\mathcal{D}$ and $zI\mathcal{D}$ are related, Lemma 2 indicates that proofs of non-$zI\mathcal{D}$ are at least as hard as the ones for non-$I\mathcal{D}$, given that to prove the former requires the construction of two models to agree on $\langle P, I \rangle$, while to prove the latter it is only required for the two models to agree on the distribution $P$.

## 3 Characterizing $zI\mathcal{D}$ Relations

The concept of confounded component (or *C-component*) was introduced in [Tian and Pearl, 2002]

to represent clusters of variables connected through bidirected edges, and was instrumental in establishing a number of conditions for ordinary identification (Def. 1). If $G$ is not a $C$-component itself, it can be uniquely partitioned into a set $\mathcal{C}(G)$ of $C$-components. We state below this definition that will also play a key role in the problem of $zI\mathcal{D}$. [3]

**Definition 3** (C-component). *Let $G$ be a causal diagram such that a subset of its bidirected arcs forms a spanning tree over all vertices in $G$. Then $G$ is a C-component (confounded component).*

A special subset of $C$-components that embraces the ancestral set of $Y$ was noted by [Shpitser and Pearl, 2006] to play an important role in deciding identifiability – this observation can also be applied to $z$-identifiability, as formulated next.

**Definition 4** (C-forest). *Let $G$ be a causal diagram, where $\mathbf{Y}$ is the maximal root set. Then $G$ is a $\mathbf{Y}$-rooted C-forest if $G$ is a C-component and all observable nodes have at most one child.*

We next introduce a structure based on C-forests that witnesses unidentifiability characterized by a pair of $C$-forests. $I\mathcal{D}$ was shown by [Shpitser and Pearl, 2006] infeasible if and only if such structure exists as an edge subgraph of the given causal diagram.

**Definition 5** (hedge). *Let $\mathbf{X}, \mathbf{Y}$ be set of variables in $G$. Let $F, F'$ be $\mathbf{R}$-rooted C-forests such that $F \cap X \neq 0$, $F' \cap X = 0$, $F' \subseteq F$, $\mathbf{R} \subset An(\mathbf{Y})_{G_{\overline{\mathbf{X}}}}$. Then $F$ and $F'$ form a hedge for $P_{\mathbf{x}}(\mathbf{Y})$ in $G$.*

The presence of this structure will prove to be an obstacle to $z$-identifiability of causal effects in various scenarios. For instance, the $p$-graph in Fig. 1(b) is a $Y$-rooted C-forest in which $P_x(y)$ will show not to be $z$-identifiable. However, different than in the $I\mathcal{D}$ case, there is no sharp boundary here, since Fig. 1(a) also contains a $Y$-rooted C-forest but $P_x(y)$ was already shown to be $zI\mathcal{D}$.

We formally show next that there is a variation of this structure that is able to capture non-$zI\mathcal{D}$ for a broad set of cases.

**Theorem 2.** *Let $\mathbf{X}, \mathbf{Y}, \mathbf{Z}$ be disjoint sets of variables and let $G$ be the causal diagram. Then, the causal effects $Q = P_{\mathbf{x}}(\mathbf{y})$ is not $zI\mathcal{D}$ if there exists a hedge $\mathcal{F} = \langle F, F' \rangle$ for $Q$ in $G_{\overline{Z}}$.*

*Proof.* The result is immediate. The existence of the hedge $\mathcal{F}$ for $Q$ in $G_{\overline{\mathbf{Z}}}$ implies that $\mathbf{Z}$ cannot help in the (ordinary) identification of $Q$. Let us assume that $Q$ is $zI\mathcal{D}$. Note that $\mathbf{Z}$ does not participate in the hedge $\mathcal{F}$ since there is no bidirected edge going towards any of its elements in $G_{\overline{\mathbf{Z}}}$, which is required by the definition of C-forest. Further, consider a parametrization such that all elements of $\mathbf{Z}$ are simply fair coins and disconnected from $\mathbf{V} \setminus \mathbf{Z}$ in $G$.

---

[3]The advent of $C$-components complements the notion of *inducing path*, which was introduced earlier in [Verma and Pearl, 1990].

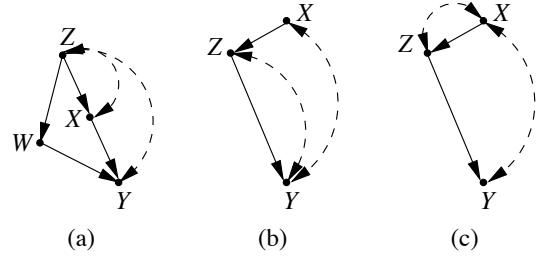

Figure 2: Graphs in which $P(y|\hat{x})$ is non-$zI\mathcal{D}$ from $do(Z)$ and there is no hedge in $G_{\overline{Z}}$.

We can now use the same proof of non-$I\mathcal{D}$ based on $\mathcal{F}$ to prove non-$zI\mathcal{D}$ in $G$. The inequality of $Q$ between the two models is obvious, and the agreement of the interventional distributions $do(\mathbf{Z})$ follows since $\mathbf{Z}$ is disconnected from $\mathbf{V} \setminus \mathbf{Z}$ by the chosen parametrization. This is a contradiction since $zI\mathcal{D}$ has to be valid for any parametrization compatible with $G$, which suffices to prove the result. □

Consider the next Corollary in regard to the $p$-graph, which is the smallest example in which $Z$ could aid in the $z$-identification of $Q$ but $Q$ is still not $z$-identifiable from $do(Z)$. This and similar structures that prevent $zI\mathcal{D}$ will be one of the base cases for our proof of completeness, which requires a demonstration that whenever the algorithm fails to $z$-identify a causal relation, the relation is indeed non-$zI\mathcal{D}$.

**Corollary 1.** *$P_x(y)$ is not $zI\mathcal{D}$ in the $p$-graph.*

*Proof.* This follows directly from Theorem 2 since there exists a hedge in $G_{\overline{Z}}$. □

The result of Theorem 2 still does not characterize the $zI\mathcal{D}$ class, which suggests that the machinery used to prove completeness in the $I\mathcal{D}$ class is not immediately applicable to the $zI\mathcal{D}$ class.

For instance, consider the graph in Fig. 2(a) (called here $bv$-graph), which does not have a hedge for $Q$ in $G_{\overline{Z}}$ but is still non-$zI\mathcal{D}$. The $bv$-graph coincides as an edge subgraph with Fig. 1(a) (note C-component induced over $\{X, Y, Z\}$), which turns out to be $zI\mathcal{D}$.

This is an interesting case, since up to this point, in ordinary identification, it was enough to locate a hedge for $Q$ as an edge subgraph of the inputted diagram, and all graphs sharing this substructure were equally unidentifiable (see Thm. 4 in [Shpitser and Pearl, 2006]) – this is no longer true here since **Z** needs to be taken into account. Mainly, note that the directed edges outside a C-component play a very critical role for the *zID* problem as the *bv*-graph demonstrates.

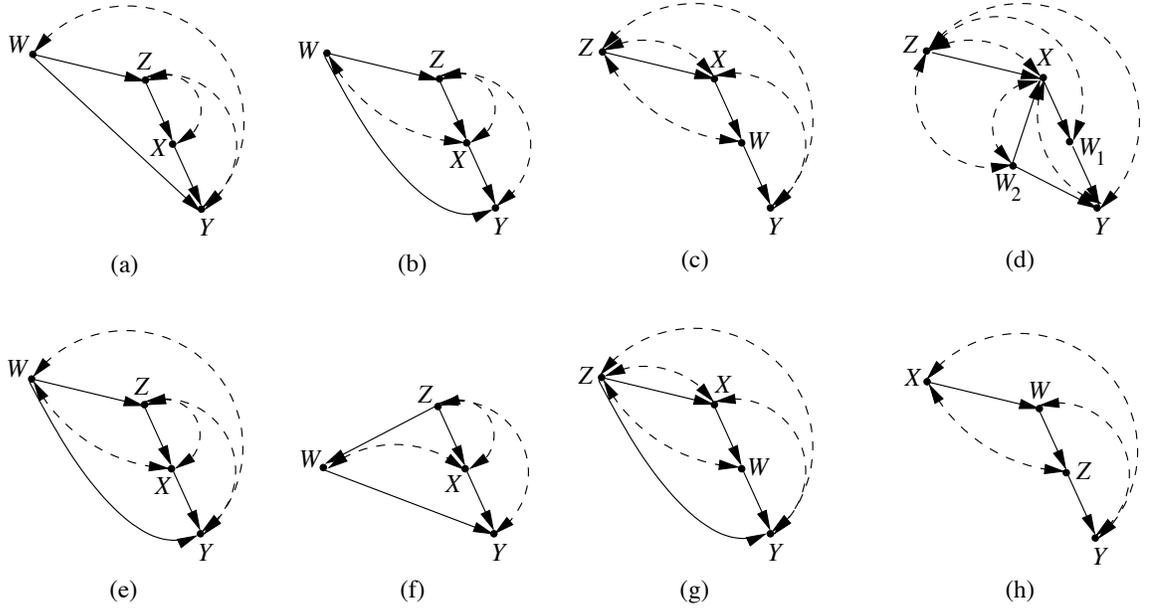

Figure 3: $P(y|\hat{x})$ is *zID* from $\langle P, do(Z) \rangle$ in the graphs in the first row (a–d), but not in the the second row (e–h).

Finally, we expand Pearl's condition [Pearl, 2000, pp. 87] in the following directions. We extend, in the intuitive way, his condition to consider when **Z** is a set of variables and, in turn, we supplement the sufficient part with its necessary counterpart. We finally have a complete characterization for the *zID* class as shown below.

**Theorem 3.** *Let* **X**, **Y**, **Z** *be disjoint sets of variables and let G be the causal diagram. The causal effect $Q = P(\mathbf{y}|do(\mathbf{x}))$ is zID in G if and only if one of the following conditions hold:*

a. *Q is identifiable in G; or,*

b. *There exists* $\mathbf{Z}' \subseteq \mathbf{Z}$ *such that the following conditions hold,*

  (i) **X** *intercepts all directed paths from* $\mathbf{Z}'$ *to* **Y**, *and*

  (ii) *Q is identifiable in* $G_{\overline{\mathbf{Z}'}}$.[4]

*Proof.* See Appendix. □

Let $Q = P(y|\hat{x})$ be the effect of interest and assume that experiments were performed over $\{Z\}$. $Q$ is *zID* from $P$ and $do(Z)$ in the graphs in Fig. 3(a-d), while they are non-*zID* in the graphs in Fig. 3(e-h). Except for the trivial case, Theorem 3 is existentially quantified and it is not immediately obvious how to efficiently select the covariates simultaneously satisfying both conditions of the Theorem. Clearly, a naive approach could lead to an exponential number of tests.

For example, consider the graph in Fig. 3(a) that is a variation of the *bv*-graph. In this graph, $Q$ is *zID* using experiments from $\{Z\}$. In turn, consider the graph in Fig. 3(e), which is the same as 3(a) but with the bidirected edge $W \leftarrow\!\!\rightarrow X$ added. Now, $Q$ is no longer *zID* for $\{Z\}$ nor $\{Z, W\}$. If we further consider the graph in Fig. 3(b) with the bidirected edge $W \leftarrow\!\!\rightarrow X$ removed from 3(e), not only $Q$ becomes *zID* for $\{Z\}$ but also for $\{Z, W\}$. This is a border case, note that if we input $\{Z, W\}$ as the surrogate variables for Pearl's criterion, it will not be able to recognize $Q$ as *zID* given the existence of the directed path $W \to Y$. Finally, if we consider the graph in Fig. 3(f) in which the directed edge $W \to Z$ is flipped from 3(b), $Q$ is no longer *zID* for neither $\{Z, W\}$ nor $\{Z\}$.

This example can be extended indefinitely but it is clear that finding a set that satisfies both conditions of the Theorem, in structures more intricate than the given 4-node example, does not follow immediately. The subject of the next section is about finding an efficient (and complete) algorithm to solve this problem.

But for now, consider the following Lemma that confirms our intuition that surrogate experiments should not disturb the causal paths (non-descendents) of the variables that are being analyzed.

---

[4] This condition can be rephrased graphically as "There exists no hedge for $Q$ as an edge subgraph in $G_{\overline{\mathbf{Z}'}}$."

**Corollary 2.** *Let $G$ be the causal diagram, $\mathbf{X}, \mathbf{Y} \subset \mathbf{V}$ be disjoint sets of variables, and $\mathbf{Z} \subseteq De(\mathbf{X})_{G_{An(\mathbf{Y})}}$. The causal effect $Q = P(\mathbf{y}|do(\mathbf{x}))$ is not zID from $P$ and $do(\mathbf{Z})$ in $G$, if $Q$ is not ID from $P$ in $G$.*

*Proof.* The result follows directly from Theorem 3. □

## 4 A Complete Algorithm for zID

In this section, we propose a simple extension of the ordinary identification algorithms to solve the problem of $z$-identifiability, which we call **ID**$^{\mathbf{z}}$ (see Fig. 4).

We build on previous analysis of identifiability given in [Pearl, 1995; Kuroki and Miyakawa, 1999; Tian and Pearl, 2002; Shpitser and Pearl, 2006; Huang and Valtorta, 2006], and we choose to start with the version provided by Shpitser (called **ID**) since the hedge structure is explicitly employed, which will show to be instrumental to prove completeness.

Before considering the technical results, we explain our strategy and how our version of the algorithm relates to the existent ones for ordinary identifiability.

**(i) $z$-identifiability (sufficiency):** Causal relations can be solved in our context through ordinary identifiability or identifiability relying on the experiments performed over $\mathbf{Z}$. The current algorithms already operate on the first part, and they proceed exploring a sequence of equalities in do-calculus based on the C-component decomposition. (The idea is to apply a divide-and-conquer strategy breaking the problem into smaller, more manageable pieces, and then to assemble them back when it is possible.) It turns out that the equalities used by the algorithm are all in the interventional space (between interventional distributions except for the base cases), which is attractive for the $zID$ problem since certain interventional distributions $\mathbf{Z}$ are already available to use.

For instance, when steps 3 or 4 succeed in their tests and, at the same time, have non-empty intersection with $\mathbf{Z}$, we exploit the common variables, updating the graph and respective data structures accordingly. We then continue solving an ordinary $ID$ instance but no longer have to identify these variables and they possibly can help in the identifiability of others.

**(ii) Non-$z$-identifiability (necessity):** The algorithm proceeds until it is not able to resolve a certain subproblem, which implies the existence of a certain hedge. Note that the given hedge can be different than the one used for $ID$ in the same graph since the experiments over $\mathbf{Z}$ possibly destroyed the original ones. Further, note that to use the given hedge to prove non-$zID$ is not immediate since, in the light of Lemma 2, more constraints need to be satisfied in order to support

function **ID**$^{\mathbf{z}}(\mathbf{y}, \mathbf{x}, \mathbf{Z}, \mathcal{I}, \mathcal{J}, P, G)$
INPUT: $\mathbf{x}, \mathbf{y}$: value assignments; $\mathbf{Z}$: variables with interventions available; $\mathcal{I}, \mathcal{J}$: see caption; $P$: current probability distribution $do(\mathcal{I}, \mathcal{J}, x)$ (observational when $\mathcal{I} = \mathcal{J} = \emptyset$); $G$: causal graph.
OUTPUT: Expression for $P_{\mathbf{x}}(\mathbf{y})$ in terms of $P, P_{\mathbf{z}}$ or **FAIL**$(F, F')$.

1  if $\mathbf{x} = \emptyset$, return $\sum_{\mathbf{v} \setminus \mathbf{y}} P(\mathbf{v})$.
2  if $\mathbf{V} \setminus An(\mathbf{Y})_G \neq \emptyset$,
    return **ID**$^{\mathbf{z}}(\mathbf{y}, \mathbf{x} \cap An(\mathbf{Y})_G, \mathbf{Z},$
        $\mathcal{I}, \mathcal{J}, \sum_{\mathbf{v} \setminus An(\mathbf{Y})_G} P, An(\mathbf{Y})_G)$.
3  Set $\mathbf{Z_w} = ((\mathbf{V} \setminus (\mathbf{X} \cup \mathcal{I} \cup \mathcal{J})) \setminus An(\mathbf{Y})_{G_{\overline{\mathbf{X} \cup \mathcal{I} \cup \mathcal{J}}}}) \cap \mathbf{Z}$.
    Set $\mathbf{W} = ((\mathbf{V} \setminus (\mathbf{X} \cup \mathcal{I} \cup \mathcal{J})) \setminus An(\mathbf{Y})_{G_{\overline{\mathbf{X} \cup \mathcal{I} \cup \mathcal{J}}}}) \setminus \mathbf{Z}$.
    if $(\mathbf{Z_w} \cup \mathbf{W}) \neq \emptyset$,
    return **ID**$^{\mathbf{z}}(\mathbf{y}, \mathbf{x} \cup \mathbf{w}, \mathbf{Z} \setminus \mathbf{Z_w}, \mathcal{I} \cup \mathbf{z_w}, \mathcal{J}, P, G)$.
4  if $\mathcal{C}(G \setminus (\mathbf{X} \cup \mathcal{I} \cup \mathcal{J})) = \{S_0, S_1, ..., S_k\}$,
    return $\sum_{\mathbf{v} \setminus \{\mathbf{y}, \mathbf{x}, \mathcal{I}\}} \prod_i$ **ID**$^{\mathbf{z}}(s_i, (\mathbf{v} \setminus s_i) \setminus \mathbf{Z},$
        $\mathbf{Z} \setminus (\mathbf{V} \setminus S_i), \mathcal{I}, \mathcal{J} \cup (\mathbf{Z} \cap (\mathbf{v} \setminus \mathbf{s_i})), P, G)$.
    if $\mathcal{C}(G \setminus (\mathbf{X} \cup \mathcal{I} \cup \mathcal{J})) = \{S\}$,
5   if $\mathcal{C}(G) = \{G\}$, **FAIL**$(G, S)$.
6   if $S \in \mathcal{C}(G)$,
    return $\sum_{s \setminus \mathbf{y}} \prod_{i|V_i \in S} P(v_i | v_G^{(i-1)} \setminus (\mathcal{I} \cup \mathcal{J}))$.
7   if $(\exists S') S \subset S' \in \mathcal{C}(G)$,
    return **ID**$^{\mathbf{z}}(\mathbf{y}, \mathbf{x} \cap S', \mathbf{Z}, \mathcal{I}, \mathcal{J},$
        $\prod_{i|V_i \in C'} P(V_i | V_G^{(i-1)} \cap S', v_G^{(i-1)} \setminus (S' \cup \mathcal{I} \cup \mathcal{J})), S')$.

Figure 4: **ID**$^{\mathbf{z}}$: Modified version of $ID$ algorithm capable of recognizing $zID$; The variables $\mathcal{I}, \mathcal{J}$ represent indices for currently active $Z$-interventions introduced respectively by steps 3 or 4. Note that $P$ is sensitive to current instantiations of $\mathcal{I}, \mathcal{J}$.

such claim. Still, it is clear that if $\mathbf{Z}$ is not involved in the hedge, it can be shown that the two problems coincide. The other cases in which $\mathbf{Z}$ has non-empty intersection with the hedge have to be handled more carefully.

Note that the key difference between **ID**$^{\mathbf{z}}$ and the original **ID** implementation is in steps 3 and 4 in which possibly some $\mathbf{Z}' \subseteq \mathbf{Z}$ is added as an interventional set, and kept as so until the end of the execution. It is clear that these additions just can represent a benefit in computing the target $Q$ since is always easier to identify a quantity in a subgraph of the original input.

We prove next soundness and completeness of **ID**$^{\mathbf{z}}$.

**Theorem 4** (soundness). *Whenever **ID**$^{\mathbf{z}}$ returns an expression for $P_{\mathbf{x}}(\mathbf{y})$, it is correct.*

*Proof.* The result is immediate since the soundness of **ID** was already established [Shpitser and Pearl, 2006, Thm. 5], which is inherited by **ID**$^{\mathbf{z}}$ by construction. Note that adding $\mathbf{Z}' \subseteq \mathbf{Z}$ as an interventional set and not trying to "identify" it later does not represent a problem, in the *zID* sense, since by assumption we can use the interventional distributions $do(\mathbf{Z})$ in the final expression returned by the procedure. □

**Theorem 5.** *Assume $\mathbf{ID^z}$ fails to z-identify $P_\mathbf{x}(\mathbf{y})$ from $P$ and $do(\mathbf{Z})$ in $G$ (executes line 5). Then there exists $\mathbf{X}' \subseteq \mathbf{X}$, $\mathbf{Y}' \subseteq \mathbf{Y}$, $\mathbf{Z}', \mathbf{Z}'' \subseteq \mathbf{Z}$ such that the graph pair $G, S$ returned by the fail condition of $\mathbf{ID^z}$ contain as edge subgraphs C-forests $F, F'$ that form a hedge for $P_{\mathbf{x}',\mathbf{z}'}(\mathbf{y}', \mathbf{z}'')$.*

*Proof.* This property is just partly inherited from the original **ID** since we can add $\mathbf{Z}' \subseteq \mathbf{Z}$ as interventional nodes along the execution of $\mathbf{ID^z}$; we also keep track of $\mathbf{Z}'' \subseteq \mathbf{Z}$ that are related to $An(\mathbf{Y})$ during the execution of the procedure (to be specified below).

Consider $G$, $\mathbf{Y_f}$, $\mathcal{I}$ and $\mathcal{J}$ local to the call in which $\mathbf{ID^z}$ exited with failure (line 5). It is true that the set $\mathbf{Y_f}$ is such that $\mathbf{Z}'' = \mathbf{Y_f} \cap \mathbf{Z}$ and $\mathbf{Y}' = \mathbf{Y_f} \cap \mathbf{Y}$. Let $\mathbf{Z}' \subseteq \mathbf{Z}$ be the active part of $\mathbf{Z}$ in the faulty call, which we kept track through $\mathcal{I} \cup \mathcal{J}$. The condition that triggered failure is that the whole graph was a single C-component. Let $\mathbf{R}$ be the root set of $G$. We can remove a set of directed arrows while keeping the root $\mathbf{R}$ such that the resulting $F$ is an $\mathbf{R}$-rooted C-forest.

Similarly to **ID**, note that since $F' = F \cap S$ is closed under descendent and only single directed arrows were removed from $S$ to obtain $F'$, $F'$ is also a C-forest. Now, $F' \cap (\mathbf{X} \cup \mathbf{Z}') = \emptyset$ and $F' \cap (\mathbf{X} \cup \mathbf{Z}') \neq \emptyset$, by construction. Also, $\mathbf{R} \subseteq An(\mathbf{Y}', \mathbf{Z}'')_{G_{\overline{\mathbf{X},\mathbf{z}'}}}$ and $\mathbf{Z}'' \subseteq An(\mathbf{Y})_{G_{\overline{\mathbf{X},\mathbf{z}'}}}$, by line 2 and 3 of the algorithm. □

**Theorem 6** (completeness). *$\mathbf{ID^z}$ is complete.*

*Proof.* By Theorem 5, $\mathbf{ID^z}$ failure implies the existence of $\mathbf{X}' \subseteq \mathbf{X}$, $\mathbf{Y}' \subseteq \mathbf{Y}$, $\mathbf{Z}', \mathbf{Z}'' \subseteq \mathbf{Z}$, and C-forests $F, F'$ that form a hedge for $P_{\mathbf{x}',\mathbf{z}'}(\mathbf{y}', \mathbf{z}'')$. Let us proceed our analysis by cases:

Case $\mathbf{Z}' = \emptyset, \mathbf{Z}'' = \emptyset$. The construction provided by [Shpitser and Pearl, 2006, Corollary 2] can be used here since this case reduces to ordinary identifiability.

Case $\mathbf{Z}' = \emptyset, \mathbf{Z}'' \neq \emptyset$. Even though $\mathbf{Z}''$ is in the root set of the hedge, and not related to the interventional part $(F \setminus F')$ where the asymmetry in the construction usually resides (to generate inequality in $Q$), the previous construction have to be used with certain caution, as given by case 1 of Thm. 3.

There is an interesting border subcase when $\mathbf{Y}' = \emptyset$. We need to keep track of $\{\mathcal{I}, \mathcal{J}\}$ since if the Z-interventions are added in step 3, we should not be concerned with summing over the assignments of the variables added, but if the Z-interventions are added in step 4, we do have to take care of this case. Note that we would have some hedge in a do-equality in the form $Q = \sum_{\mathbf{z}''} P_{\mathbf{x}'}(\mathbf{z}'') f(\mathbf{x}, \mathbf{y}, ...)$, in which if $f(.)$ is identifiable and uniformly distributed, $Q$ would equate in both models and spoil the counter-example. The problem is not difficult to fix, and we just have to create a map for $f()$ that is non-uniform. (See Thm. 3.)

Case $\mathbf{Z}' \neq \emptyset, \mathbf{Z}'' = \emptyset$. The construction provided in cases 2 and 3 of Thm. 3 were more involved since it was not know a priori which C-factor yielded the "faulty" call. In the $\mathbf{ID^z}$ case, we already located the hedge based on the trace of the algorithm, then we can essentially use the same construction of these cases to provide a counterexample.

Case $\mathbf{Z}' \neq \emptyset, \mathbf{Z}'' \neq \emptyset$. The construction provided in the two previous cases are not incompatible, and they can be combined to provide a counter-example to this scenario.

Moreover, the previous constructions were given over the subgraph $H$ of $G$, and how to extend the counter-example to $G$ is discussed in Theorem 3. □

**Corollary 3.** *The rules of do-calculus, together with standard probability manipulations are complete for determining z-identifiability of $P_\mathbf{x}(\mathbf{y})$.*

*Proof.* It was already shown [Shpitser and Pearl, 2006, Thm. 7] that the operations of **ID** correspond to sequences of standard probability manipulations and application of the rules of do-calculus, which is also true by construction for $\mathbf{ID^z}$, and so the result follows. □

## Conclusion

This paper was concerned with a variation of the identifiability problem in which experiments can be conducted over a subset of the variables $\mathbf{Z}$ in addition to the assumptions embodied in a causal digram $G$ and the statistical knowledge given as a probability distribution. (If $\mathbf{Z}$ is an empty set, the two problems coincide.)

We provide a graphical necessary and sufficient condition for the cases when the causal effect of an arbitrary set of variables on another arbitrary set can be determined uniquely from the available information. We further provide a complete algorithm for computing the resulting mapping, that is, a formula fusing available observational and experimental data to synthesize an estimate of the desired causal effects. Furthermore, we use our results to prove completeness of *do*-calculus in respect to the *z*-identifiability class.

Our results were developed in a non-parametric setting in the tradition of the do-calculus. For a future research direction, it would be interesting to explore how experimental data can aid the identification in the linear case. This is a harder problem, since a complete characterization of ordinary identifiability (i.e., $\mathbf{Z} = \emptyset$) in the linear case is still an open problem.

This paper complements two recent works on generalizability of causal and statistical knowledge. The first, dubbed "transportability" [Pearl and Bareinboim, 2011; Bareinboim and Pearl, 2012b] , deals with transferring causal information from an experimental to an observational environment, potentially different from the first. The second, called "selection bias" [Bareinboim and Pearl, 2012c], deals with extrapolation between an environment in which samples are selected preferentially and one in which no preferential sampling takes place. The extrapolation involved in z-Identification problems takes place between two different regimes; one in which experiments are performed over $\mathbf{Z}$, and one in which future experiments are anticipated over $\mathbf{X}$. Extensions to "meta synthesis" tasks, where information from multiple heterogeneous sources are combined to increase the effective sample size, are considered in [Pearl, 2012b; 2012a].

## Acknowledgement

The authors would like to thank the reviewers for their comments that help improve the manuscript. This research was supported in parts by grants from NIH #1R01 LM009961-01, NSF #IIS-0914211 and #IIS-1018922, and ONR #N000-14-09-1-0665 and #N00014-10-1-0933.